# ASPECT-BASED SENTIMENTAL ANALYSIS FOR TRAVELLERS' REVIEWS

Mohammed Saad M Alaydaa, Jun Li, Karl Jinkins

School of Aerospace, Transport and Manufacturing (SATM), Cranfield University,
Cranfield MK43 0AL, Bedfordshire, UK
`{m.s.alanazi, jun.li, k.w.jenkins}cranfield.ac.uk`

## ABSTRACT

*Airport service quality evaluation is commonly found on social media, including Google Maps. This valuable for airport management in order to enhance the quality of services provided. However; prior studies either provide general review for topics discussed by travellers or provide sentimental value to tag the entire review without specifically mentioning the airport service that is behind such value. Accordingly; this work proposes using aspect based sentimental analysis in order to provide more detailed analysis for travellers' reviews. This works applied aspect based sentimental analysis on data collected from Google Map about Dubai and Doha airports. The results provide tangible reasons to use aspect based sentimental analysis in order to understand more the travellers and spot airport services that are in need for improvement.*

## 1. INTRODUCTION

The reviews provided by travellers hold immense significance for the aviation industry. These reviews have the potential to strongly influence travellers' decision when it comes to selecting an airport [1-5]. Even minor improvements in airport services can lead to positive changes in travellers' perceptions and enhance their overall airport experience [6-9]. Moreover; travellers' positive-sentiment is considered among the competitive features of airports [9]. Given that travellers can easily access and refer to other travellers' online reviews, airport management must prioritise the Airport Service Quality (ASQ). In order to understand the key areas that airport management should focus on to enhance positive reviews, researchers have developed tools [1] for extracting and analysing travellers' reviews.

This study employed aspect-based sentimental analysis in order to explicitly tag every airport service-mentioned in the traveller's feedback with positive/negative. This; in contrast to prior studies that tag the entire review with positive/negative; clearly assists airport management to spot services that are in need for improvement.

## 2. LITERATURE REVIEW

Most of the studies, particularly those that use secondary data such as Twitter, Google Review, airline quality or Skytrax employ topic modelling and sentimental analysis. The rest mostly use statistical analysis based on primary dataset collected directly from travellers to investigate the impact of ASQ on travellers' satisfaction, revisits and reviews. Dhini and Kusumaningrum [4] used Support vector Machine (SVM) and Naïve Bayes Classifier to classify traveller's feedback

(reviews from google) into positive or negative. Meanwhile; 20,288 online reviews posted between 2005 and 2018 on TripAdvisor were analysed by Moro, et al. [10], Heat maps for airport hotel services and sentimental status of the guests were reported. Deep learning networks (CNN and LSTM) were developed by Barakat, et al. [3] to recognize sentimental status (positive/negative) of travellers' posts found in US Airline Sentiment dataset (14487 records) and AraSenTi dataset (15,752 records). Similar work on twitter dataset (London Heathrow airport's Twitter account - dataset includes 4,392 tweets) by Martin-Domingo, et al. [11], topic modelling was used. Online reviews platform from Skytrax (2,278) were investigated by Kiliç and Çadirci [8], Halpern and Mwesiumo [12] via multinomial logit model, topic modelling, sentimental analysis, and emotion recognition to spot airport services that receive high positive/negative feedback. [13] used topic modelling and sentimental analysis with 42,137 reviews collected from Google Maps. Shadiyar, et al. [14], Bunchongchit and Wattanacharoensil [15] employed several methods (text mining analysis, semantic network analysis, frequency analysis and linear regression analysis) to assess 1,693 and 7,358 reviews related to airport and flights respectively.

In effort to determine a list (scale) of airport services that could be used with online social media posts- in analogy with airport service quality scale developed from survey; Tian, et al. [16] used text mining and sentiment analysis to come up with scale of 6 airport service scale. Topic modelling is a popular method used to analyse online comments made by travellers, with tools such as Latent Dirichlet Allocation (LDA) commonly employed to investigate the major airport services [8, 13]. These tools combined with dimension reduction approaches are applied to identify a limited number of topics from the comments of the travellers.

The above discussion was extensively used NLP qualitative approach such as Latent Semantic Analysis (LSA) and Latent Dirichlet Allocation (LDA) to determine airport services mentioned in travellers' feedback (via topic modelling). Later, sentimental analysis used to know positivity/negativity of travellers' feedback. Then, recognizing sentiment scores (i.e., positive and negative) [8] or sentimental values (i.e., positive, negative, and neutral) are not sufficient to accurately reveal people's specific sentiments [16] or which specific airport service targeted by this feedback. Lee and Yu [13] applied LDA to predict the star ratings of airports from sentimental scores. Bae and Chi [17] employed an alternative approach called content analysis to distinguish between satisfied and dissatisfied travellers using their online reviews. The study found that dissatisfied travellers frequently used words such as "security," "check," "staff," "flight," and "line," whereas satisfied travellers often used words like "staff," "terminal," "time," "clean," "immigration," and "free."

In recent years, Machine Learning, especially supervised methods such as Deep Learning, have gained popularity in predicting traveller sentimental values. Li, et al. [2] reported that studies using social media data to predict sentimental values based on Vader and LSVA. Taecharungroj and Mathayomchan [18] found that the quality of airport services can be measured by sentimental values associated with various services, such as access, check-in/security, wayfinding, facilities, airport environment, and staff. Barakat, et al. [3] used thousands of English and Arabic tweets to train CNN and LSTM models to predict positive or negative traveller sentiments toward airport services. Although the LSTM model showed better prediction, the difference is insignificant. Kamış and Goularas [19] evaluated several Deep Learning architectures with different datasets and found that the best performance was achieved when LSTM and CNN were combined. Generally, studies in Machine Learning and Deep Learning on airport service quality and travellers' sentimental value since 2018 have been limited.

## 2.1. Airport Services

Despite the various techniques used to measure ASQ, most studies come to a similar conclusion that certain airport services are more likely to receive positive reviews if they are effectively managed. However, there is no standardized way of listing the airport services that should be focused on. Some researchers, like Gajewicz, et al. [6], evaluate facility attributes such as waiting time, cleanliness, efficiency, and availability of services individually, while others consider these attributes as a whole. Additionally, some researchers use broader terms, such as facilities to include amenities like food, restaurants, and ATMs, while others are more specific. Consequently, different lists of airport services are found in the studies, making it difficult to standardize a list of services in airports to be evaluated. Table 1 provides a list of airport services that cover all explicit facilities in the airport, based on the review.

Table 1. A list of airport services and specification

| Services | Specification |
|---|---|
| Access | Transportation, parking facilities, trolleys, baggage, and cars etc. |
| Check-in and security | Waiting time, check-in queue/ line, efficiency of check-in staff, and waiting time at security inspection etc. |
| Facilities | ATM, toilets, and restaurants etc. |
| Wayfinding | Ease of finding your way through airport, and flight information screens etc. |
| Airport environment | Cleanliness of airport terminal, ambience of the airport, etc. |

Table 2 provides a list of airport services based on a sample of 13 studies, where check-in is mentioned most frequently and queuing/waiting time occurs in studies least frequently. However, there are some inconsistencies in how certain services are categorized. For example, some studies treat check-in and security as a single category, while queuing/waiting time is classified as a feature for arrival. Additionally, some airport services are uniquely featured in specific studies, such as services cap [20], prime services [21], and airport appearance [22].

Table 2. Airport services reported in the studies

| Airport service | Study |
|---|---|
| Passport control, arrival services, airport environment, wayfinding, airport facilities, check-in, security, and access | [23] |
| Access, facilities, wayfinding, environment, personnel, check-in, security, and arrival | [2] |
| Access, check-in, passport, wayfinding, facilities, environment, arrival, people (personnel), and waiting | [3] |
| Signage and wayfinding, information, security, waiting times, staff, cleanness, comfort, and availability/efficiency of the airport services | [6] |
| Access, Security, check-in, facilities, wayfinding, environment, and arrival | [24] |
| Airport staff and queuing times | [12] |
| Security, check-in, wayfinding, environment, access, arrival services and airport facilities | [24] |
| Facilities, check-in, services cap, security and ambience | [20] |
| Traffic, check-in, signs and wayfinding, environment, security and passport/ID card inspection, entry procedures, and facilities | Liu and Zheng [25] |
| Non-processing (main facilities, value addition) and processing (queue and waiting time, staff (helpfulness and communication), prime services) | [21] |
| Seat comfort, staff, food and beverage, entertainment, ground services, and value for money | [14] |
| Access, check-in/security, way finding, facilities, environment, and staff | [3] |
| Services, airport appearance, check in/out services, and waiting time | [22] |

The gap found in the current studies that employed sentimental analysis was given a polarity value (negative, positive or neutral) as the overall sentimental value for travellers' feedback. in many cases, travellers feedback contains several sentimental values (positive, negative or neutral) tagging different airport services. Therefore, tagging the feedback with a single sentimental value may underestimate other important values that could alert airport management to drawbacks in the services provided. Li, et al. [2] found a significant relationship between review rating and some specific airport services mentioned in google map reviews. Accordingly, Aspect-Based Sentimental Analysis (ABSA) could provide more detailed information about the sentimental values that travellers want to deliver.

## 3. METHODOLOGY

### 3.1. Datasets

Datasets were collected from Google Maps, and the tool provided by outscraper.com was used to collect the data of two famous airports in the Arabic peninsula: Dubai and Doha. The number of reviews collected related to Doha and Dubai airports were 11400 and 16170, respectively. No specific dates were set, but most of the reviews were done during the Covid-19 outbreak. The data items used were travellers' reviews and review rate (1~5)- if the rating is greater than or equal to 3, the polarity is positive; otherwise, it is negative [26]. Other items were removed because they either revealed the personal information about the reviewer (name, image, etc.) or were related to the time and date.

### 3.2. Method

To ensure a proper aspect-based sentimental analysis, the entire review is divided into a set of sentences using NLTK Tokenizer. This set of sentences is fed to a method that use *textblob library* to correct misspellings. The list of aspects extracted from table 2 ("access", "security", "check-in", "facilities", "Wayfinding", "arrival", "staff", "terminal") and related terms (e.g., facilities: food, seats, toilet wifi etc.) will be searched inside the sentence. If any aspect exists, the sentence will be fed to an Aspect-based library (deberta-v3-base-absa-v1.1). The highest score (positive or negative) that comes out of the Aspect-based will tag the sentence. Eventually; a matrix of aspects will be the output that reveal the polarity of traveller's feedback about all possible airport services mentioned in the feedback.

## 4. RESULTS AND DISCUSSION

A sample of output is presented in table 3. It is important to mention that the number under Facilities column in the first row is the average value of the values predicted by aspect model for sentimental values of "toilets, shops, seats" because all those terms are under facilities aspect. Similarly in the second row, but this time with negative values. Zero values means that the aspects are missing in the travellers' feedback. keywords column refers to services that explicitly mentioned in the traveller's feedback. This to give an idea for which; for instance; facility exactly the traveller gave feedback about. Moreover; this is in line with prior studies that tried to shorten the list of services that could be tracked in feedback in order to give a more accurate result to airport management. Moreover; the results could be averaged in order to provide overall sentimental value for each airport service. This will provide more quantitative data instead of providing a list of words that frequently used under an estimated topic-as LSA and LDA do.
In essence; utilizing aspect-based sentimental analysis provides more details regarding the sentimental values exist in the travellers' reviews, in contrast to one single value that tagged the entire review. Moreover; in contrast to LSA and LDA, aspect-based should be used while the aspects (in this context is airport services) already known; meanwhile, LDA needs in advanced the possible topics number in order to come up with a group of words that could describe specific topic.
On other hand; the current Aspect-based sentimental analysis tools were trained on data related to restaurant, mobile phones and computer sells, which explains the average accuracy the tool that used in this work. The accuracy was ~80%. It was noticed that when airport services repeated several times in the feedback, it makes the predicted sentimental value as a collective value, which may not reflect the real sentimental value. Therefore, there is a need to train the current aspect-based tools with datasets related to travellers' feedback.

Table 3. a sample of the output

| Feedback | Keywords | F | T | A | S | C | W | A | ST |
|---|---|---|---|---|---|---|---|---|---|
| Terminal Its clean and good for shopping if you like designer clothing shops, watches, perfume Limited selection of sunglasses, from each brand all in one shop Only water fountain to refill your water bottle Some nice comfy lounging style seats can be found if you look hard enough A good area to get some work done with charging points and desks are also here Get a full body massage for about £. Toilets are decent | Terminal, seats, toilets, shops | .94 | .98 | 0 | 0 | 0 | 0 | 0 | 0 |
| Could be delivered better amenities for the passengers The wifi signal is really slow and weak, can't even make a WhatsApp call Food in restaurants in the airport is way too expensive and bland Customer service, cleanliness and the shuttle service between terminals are decent Strollers are provided for the kids free of charge inside the airport is helpful | Amenities, wifi, food, shuttle | -.89 | .99 | 0 | 0 | 0 | 0 | 0 | 0 |

Note: F-facilities; T-terminal; A-access; S-security; Check-in; Wayfinding; A-Arrival; ST-Staff

## 5. CONCLUSION

NLP provides qualitative approaches such as LSA and LDA in order to collectively bring together frequent words that possibly shape a topic. However; the results need human intervention in order to understand the topic and sentimental values related. The other group of studies concerned to know the polarity of the entire review; which makes it difficult to know the airport service that is specifically behind such polarity. In contrast, this work found that aspect-based sentimental analysis can deliver more accurate answer regarding the polarity of every airport service mentioned in the travellers' reviews. Yet; Aspect-based sentimental analysis needs an explicit list of aspects in order to predict the sentimental value. Accordingly; and based on prior studies, this study found 8 airport services that frequently reported at prior studies (table 2) and been used here to deliver more accurate sentimental values that can help airport management to spot services that travellers complain more about. This work reports part of project, which is in progress to develop multi-label model to predict the airport services and their sentimental values.

A more work is needed to develop a specialized aspect-based that consider aspects related to airport services. This can be done by retrained the current tools with datasets similar to the ones collected in this study. The future work is to come up with tool that is capable to located airport services and its polarity within traveller's feedback.

## AUTHORS

Mohammed Saad M Alaydaa is a PhD candidate at School of Aerospace, Transport and Manufacturing (SATM), Cranfield, UK. Currently; He works on NLP with passengers' feedback in order to develop Aspect-based Sentimental analysis for airport services.

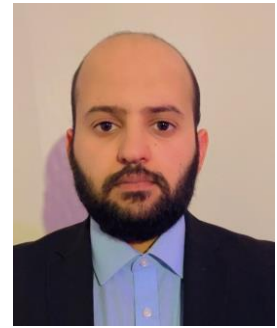

Jun Li, has a BSc, MSc and PhD in Computer Science, Software Engineering and Neural Networks respectively, obtained from QingDao University and LondonMet University. His teaching and research expertise are in the areas of Data Analytics, Machine Learning, Computer Science and Mathematical Modelling applied to various domains.

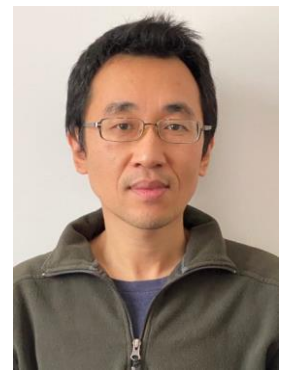

Karl Jinkins, gained a PhD from the University of Manchester which focused on computational and experimental water waves breaking interacting with coastal structures.
Professor Jenkins has also worked in industry for Allot and Lomax Consulting Engineers and Davy Distington Ltd, working on various commercial CFD codes and training engineers in their use. He has worked on adaptive parallel grid techniques, and has developed parallel codes for academic use and for blue chip companies such as Rolls Royce plc

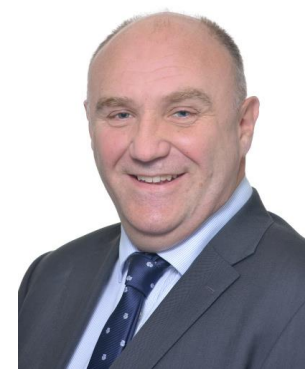